\documentclass{INTERSPEECH2023}

\interspeechcameraready

\title{Can Contextual Biasing Remain Effective with Whisper and GPT-2?}
\name{Guangzhi Sun, Xianrui Zheng, Chao Zhang, Philip C. Woodland\thanks{Guangzhi Sun and Xianrui Zheng are supported by Cambridge Trust and an Amazon Studentship respectively. This work has been performed using resources provided by the Cambridge Tier-2 system operated by the University of Cambridge Research Computing Service (www.hpc.cam.ac.uk) funded by EPSRC Tier-2 capital grant EP/T022159/1.}}
\address{
  University of Cambridge Department of Engineering, Trumpington St., Cambridge, CB2 1PZ U.K.}
\email{\texttt{\{gs534,xz396,cz277,pcw\}@eng.cam.ac.uk}}
\begin{document}

\maketitle
\ninept
 
\begin{abstract}
%End-to-end automatic speech recognition (ASR) and large language models (LLMs) have been scaled up using vast amounts of training data in recent years. Despite the coverage of the training data, infrequent content words may still exhibit suboptimal ASR performance, prompting considerations of contextual biasing as a viable remedy for this issue. To this end, this paper discussed the effectiveness of neural contextual biasing for Whisper combined with GPT-2. Specifically, this paper proposed to integrate an adapted tree-constrained pointer generator (TCPGen) component for Whisper and a dedicated training scheme without the need for Whisper fine-tuning. Experiments across three datasets showed a large reduction in biasing words with a biasing list of 1000 words. Contextual biasing was more effective when applied to domain-specific data and can boost the performance of universal models without losing their generalisability.
End-to-end automatic speech recognition (ASR) and large language models, such as Whisper and GPT-2, have recently been scaled to use vast amounts of training data. Despite the large amount of training data, infrequent content words that occur in a particular task may still exhibit poor ASR performance, with contextual biasing a possible remedy. This paper investigates the effectiveness of neural contextual biasing for Whisper combined with GPT-2. Specifically, this paper proposes integrating an adapted tree-constrained pointer generator (TCPGen) component for Whisper and a dedicated training scheme to dynamically adjust the final output without modifying any Whisper model parameters. Experiments across three datasets show a considerable reduction in errors on biasing words with a biasing list of 1000 words. Contextual biasing was more effective when applied to domain-specific data and can boost the performance of Whisper and GPT-2 without losing their generality.

\end{abstract}

\section{Introduction}

Recent advancements in software and hardware have facilitated the significant expansion of end-to-end automatic speech recognition (ASR) systems through the use of copious amounts of training data. In this regard, large-scale supervised models with extensive training, such as \textit{Whisper} \cite{whisper}, have been developed in addition to numerous foundation models featuring self-supervised training, e.g. wav2vec2.0 and GPT-2 \cite{whisper,GPT-2}. {Whisper} contains a series of ASR models that have been trained in a supervised manner on hundreds of thousands of hours of labelled speech, which can be broadly applied across a diverse range of ASR tasks. Unlike self-supervised foundation models, Whisper aimed to provide a readily deployable ``out-of-the-box" solution for ASR tasks in various environments, circumventing the need for supervised fine-tuning of a decoder for every deployment. This approach aims to mitigate the issue of decoder brittleness when fine-tuned for specific tasks.

Notwithstanding the comprehensive training data used in Whisper and GPT-2, domain-specific words, such as proper nouns, are rare or absent in the general training data and may be susceptible to errors when used for a specific domain. On the other hand, fine-tuning these models on a limited amount of domain-specific data may potentially compromise their ability to generalise to other tasks. Therefore, it is essential to enhance the performance of these models on such domain-specific words without sacrificing their capacity for generalisation, and contextual biasing is one possible solution. Contextual biasing aims to incorporate contextual knowledge into end-to-end ASR systems. Contextual knowledge is often represented as a biasing list, comprising words that carry important information and are essential to downstream tasks, such as restaurant names in an ontology for a task-oriented dialogue system. The inclusion of these words in the biasing list has been shown to significantly improve recognition performance.

Based on the aforementioned concerns and the solution, this paper endeavours to answer the question: ``Can contextual biasing remain effective in Whisper and GPT-2?" with the constraint that the parameters of both models are unchanged. Specifically, TCPGen is chosen as the neural biasing component as it achieves biasing via \textit{distribution-level adaptation} which does not require any modifications to the Whisper model. This paper proposes modifications TCPGen to be compatible with Whisper, together with a dedicated training scheme on small-scale \textit{task-specific} data. Notably, the proposed method enabled the portability of TCPGen as a neural biasing component, allowing it to maintain the "out-of-the-box" nature of Whisper without the necessity of fine-tuning any of its components\footnote{Code is available at {https://github.com/BriansIDP/WhisperBiasing.git}}.

% Specifically, Whisper was used as an exemplary system of large-scale universal ASR models, together with a pre-trained GPT-2 representing large-scale LMs. The combination of Whisper and GPT-2 adopted a rescoring scheme with an internal LM estimation (ILME) technique. 
% For neural biasing, this paper adopted the tree-constrained pointer generator (TCPGen) as an example neural contextual biasing component, modified to be compatible with Transformer-based decoders. In addition, a dedicated training scheme for TCPGen was introduced, which aimed to adapt TCPGen with small-scale \textit{task-specific} data, in addition to the modified structure that rendered TCPGen compatible with Whisper. 
% The adapted TCPGen and the training scheme were collectively referred to as Whisper Biasing (\textit{WB}) in this paper. 
% Notably, the proposed WB enabled the portability of TCPGen as a neural biasing component, allowing it to maintain the "out-of-the-box" nature of Whisper without the necessity of fine-tuning any of its components.

% As a neural-based biasing method, TCPGen constructs a neural shortcut that directly connects the biasing list to the ASR model output via a pointer generator without the need for any modifications to the model's original architecture. Moreover, TCPGen structures the biasing list using a word-piece-level prefix tree, enabling it to handle biasing lists of thousands of words efficiently.

Experiments conducted across LibriSpeech, SLURP and Dialogue State Tracking Challenge (DSTC2) datasets demonstrated that TCPGen consistently improved the performance of biasing words with a biasing list of 1000 biasing words. 
% Experiments were conducted across three different data sets, including LibriSpeech audiobook data as the set being evaluated in \cite{whisper}, as well as the SLURP and Dialogue State Tracking Challenge (DSTC2) data. The result demonstrated that TCPGen consistently improved the Word Error Rate (WER) across all sets, particularly for biasing words.  
Contextual biasing was found to be more effective when applied to \textit{domain-specific} data such as SLURP and DSTC2, especially when using a more focused biasing list.
% With a biasing list of 1000 words, contextual biasing achieved around 20\% WER improvements on biasing words for \textit{generic} data, i.e. LibriSpeech, and much larger improvements for domain-specific data including SLURP and DSTC2.

The rest of this paper is organised as follows: Section \ref{sec:rel} summarises related work. Section \ref{sec:method} explains the use of TCPGen in Whisper and its training scheme. Section \ref{sec:exp} describes the experimental setup, followed by Section \ref{sec:result} analysing results. The paper concludes in Section \ref{sec:con}.

\section{Related Work}
\label{sec:rel}
\subsection{Whisper and GPT-2 for ASR}
% \vspace{-0.2cm}
% \subsubsection{Whisper}

\textbf{Whisper} \cite{radford2022robust} is a Transformer-based \cite{transformer} encoder-decoder model trained purely on supervised data for multiple languages and multiple speech-related tasks. The training data contained paired multilingual speech and text obtained from the Internet. The amount of training data consisted of 680k hours of audio, of which 117k hours were non-English data. The released Whisper models have various sizes, ranging from 39M parameters (4 encoder blocks/4 decoder blocks) to 1550M parameters (32 encoder blocks/32 decoder blocks). 
The tokeniser used the same set of Byte-Pair Encoding (BPE) tokens as the GPT-2 \cite{GPT-2} model for English-only models and with extra tokens for multilingual models.
Whisper provides competitive ASR results across multiple languages and domains, without further fine-tuning.  

\textbf{GPT-2} \cite{GPT-2} is a Transformer based language model trained to predict the next wordpiece, without explicit supervision. 
The training data were gathered from a social media platform, where the resulting text was more than ten billion words. 
GPT-2 and its predecessor GPT \cite{gpt} have been used to improve the performance of an existing ASR model \cite{emptrans,adaptgptbert}.

\subsection{Contextual biasing}
Contextual biasing has typically used shallow fusion with a weighted finite-state transducer \cite{shallow_context_1,shallow_context_3}, via deep biasing \cite{deep_context_1,deep_context_2,deep_context_3,deep_context_4,deep_context_5}, or via deliberation models \cite{delib}. However, recently, combinations of shallow fusion and deep biasing \cite{deepshallow, DBRNNT} or neural shortcut-based methods \cite{MEM, TCPGen} have been proposed. The work in \cite{DBRNNT} proposed and validated a simulation of contextual biasing on open-source data, which was also adopted in subsequent work such as \cite{TCPGen}. Unlike deep biasing methods which required modifications to the ASR model structure to take additional inputs, neural shortcut-based methods directly modified the network output distribution without changing the structure of the ASR model. 

In particular, \cite{MEM} and \cite{TCPGen} both used the neural shortcut between the biasing list and the final output. In contrast to \cite{MEM} which used a Transformer context encoder, TCPGen \cite{TCPGen,tcpgengnn} is a light-weight component which structured biasing lists as wordpiece prefix trees to achieve efficient processing of biasing lists containing thousands of words. Thus, TCPGen was selected as the biasing component for Whisper to achieve efficient biasing without fine-tuning any Whisper parameters.

\section{Methodology}
\label{sec:method}
\subsection{Tree-constrained pointer generator}

TCPGen is a neural network-based component combining the symbolic prefix-tree search with a neural pointer generator \cite{pointer_1} for contextual biasing, which enables end-to-end optimisation with ASR systems. At each output step, TCPGen calculates a distribution over all valid wordpieces constrained by a word-piece-level prefix tree built from the biasing list (referred to as the TCPGen distribution). TCPGen also predicts a generation probability indicating how much contextual biasing is needed at a specific step. The final output is the interpolation between the TCPGen distribution and the original ASR model output distribution, weighted by the generation probability.

Specifically, a set of valid wordpieces, $\mathcal{Y}^\text{tree}_i$, is obtained by searching the prefix tree with a given history. Then, denoting $\mathbf{x}_{1:T}$ and $y_i$ as input acoustic features and output wordpieces respectively, $\mathbf{q}_i$ as the query vector carrying history and acoustic information, and $\mathbf{K}=[...,\mathbf{k}_j,...]$ as the key vectors, scaled dot-product attention is performed to compute the TCPGen distribution $P^\text{ptr}$ and an output vector $\mathbf{h}^{\text{ptr}}_i$ as shown in Eqns. \eqref{eq:TCPGen_attention} and \eqref{eq:TCPGen_value}.
\vspace{-0.2cm}
\begin{equation}
    P^{\text{ptr}}(y_{i}|y_{1:i-1},\mathbf{x}_{1:T}) = \text{Softmax}(\text{Mask}(\mathbf{q}_i\mathbf{K}^\text{T}/\sqrt{d}))
    \label{eq:TCPGen_attention}
    \vspace{-0.1cm}
\end{equation}
\begin{equation}
    \mathbf{h}^{\text{ptr}}_i = \sum\nolimits_{j} P^{\text{ptr}}(y_i=j|y_{1:i-1},\mathbf{x}_{1:T})\,\mathbf{v}^\text{T}_j
    \label{eq:TCPGen_value}
    \vspace{-0.1cm}
\end{equation}
where $d$ is the size of $\mathbf{q}_i$ (see \cite{transformer}), Mask$(\cdot)$ sets the probabilities of wordpieces that are not in $\mathcal{Y}^{\text{tree}}_i$ to zero, and $\mathbf{v}_j$ is the value vector relevant to $j$. This paper specifically focuses on the attention-based encoder-decoder (AED) ASR model. In AED, the query combines the context vector and the previously decoded token embedding, while the keys and values are computed from the decoder wordpiece embedding, with a shared projection matrix. The generation probability which takes a value between 0 and 1, is calculated using the decoder hidden state and the TCPGen output vector $\mathbf{h}^{\text{ptr}}_i$. Then, the final output can be calculated as shown in Eqn. \eqref{eq:TCPGen_final}.
\vspace{-0.1cm}
\begin{equation}
    P(y_i) = P^{\text{mdl}}(y_i)(1-{P}^\text{gen}_i) + P^{\text{ptr}}(y_i)P^\text{gen}_i
    \label{eq:TCPGen_final}
    \vspace{-0.1cm}
\end{equation}
where conditions, $y_{1:i-1}, \mathbf{x}_{1:T}$, are omitted for clarity. $P^{\text{mdl}}(y_i)$ represents the output distribution from the standard end-to-end model, and $P^{\text{gen}}_i$ is the generation probability. 

\subsection{Biasing Whisper with TCPGen}

\begin{figure}[t]
    \vspace{-0.3cm}
    \centering
    \includegraphics[scale=0.26]{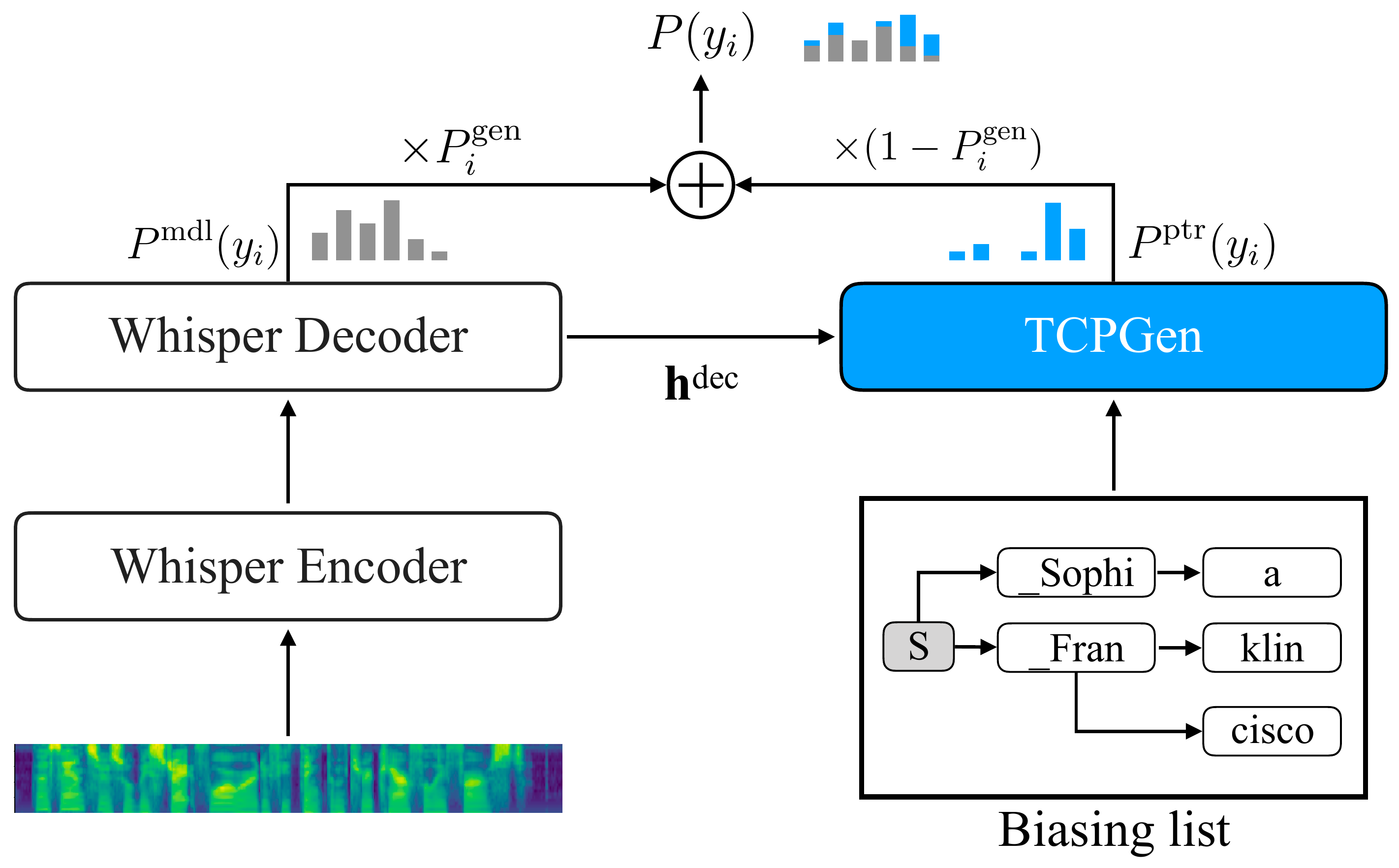}
    \vspace{-0.2cm}
    \caption{Integration of TCPGen in Whisper with corresponding terms in Eqn. (\eqref{eq:TCPGen_final}). The $\bigoplus$ symbol represents linear interpolation with weights $P^\text{gen}$ predicted by TCPGen. Only TCPGen part is updated during training. Biasing list example here contains \texttt{Sophia}, \texttt{Franklin} and \texttt{Francisco}, and underscore marks the word start}
    \vspace{-0.3cm}
    \label{fig:whisperbaising}
\end{figure}
\vspace{-0.1cm}

The integration of TCPGen into the Whisper model is illustrated in Fig. \ref{fig:whisperbaising}. TCPGen takes as input the Whisper final decoder block hidden state $\mathbf{h}^\text{dec}$ to derive the query to calculate attention as shown in Eqn. \eqref{eq:tcpgenwhisper_1}
\vspace{-0.1cm}
\begin{equation}
    \mathbf{q}_i = \text{ReLU}(W\mathbf{h}_i^\text{dec})
    \label{eq:tcpgenwhisper_1}
\end{equation}
% \vspace{-0.1cm}
where $W$ are the parameter matrix. Key and value vectors directly used the decoder wordpiece embeddings. The computation of the generation probability also used the final hidden state of the decoder, as shown in Eqn. \eqref{eq:tcpgenwhisper_2}.
% \vspace{-0.1cm}
\begin{equation}
    P^\text{gen}_i = \sigma(W_1\mathbf{h}_i^\text{dec} + W_2 \mathbf{h}_i^\text{ptr})
    \label{eq:tcpgenwhisper_2}
\end{equation}
% \vspace{-0.1cm}
where $W_1$ and $W_2$ are the parameter matrices. Then, through Eqn. \eqref{eq:TCPGen_final}, TCPGen leveraged this distribution-level adaptation to bias the Whisper output, by only requiring the output from the final decoder block of Whisper. During training on task-specific data, the entire Whisper model is frozen and only parameters in TCPGen are updated. This way, Whisper can still generalise to other data while TCPGen as a biasing component provides task-specific biasing information. Moreover, as the word frequencies of the training data for Whisper are not available, a word-error-based approach was adopted to extract a good biasing list for training instead of a frequency-based one. This was achieved by decoding the task-specific training set and gathering distinct word tokens with errors higher than average.

\subsection{Rescoring with GPT-2}

GPT-2, as an off-the-shelf large pretrained LM, is used to rescore the $N$-best hypotheses from the ASR model, as the Whisper large model has a different set of wordpiece units to the GPT-2 model. To improve the effectiveness of using GPT-2, an ILME method was also used following \cite{ilme}.
The $N$-best hypotheses were derived by the beam search algorithm during Whisper decoding, together with the log probability scores associated with each hypothesis.
% with associated \textbf{Whiper\_Score}, which is the log probabilities from the Whiper model. 
In addition to GPT-2, the log probability of each hypothesis subtracts the internal LM score, which is estimated by forwarding the Whisper model with the encoder output set to all-zero vectors.
% To find the ILM score, only the decoder part of the Whisper model was used. 
% The encoder features to the decoder are zeros and the decoder takes each of the $N$-best hypotheses to compute the log probability for that hypothesis. 
The total score of each hypothesis based on which the re-ranking is performed is computed by:
\begin{align}
\hspace{-0.2cm}\text{log}P_\text{Whisper}(Y|X) - \lambda_\text{ilm} \text{log}P_\text{ilm}(Y) + \lambda_\text{GPT-2} \text{log}P_\text{GPT-2}(Y)
% \textbf{Whiper\_Score} - \lambda_1 \textbf{ILM\_Score} + \lambda_2 \textbf{GPT-2\_Score}
\end{align}
% where \textbf{GPT-2\_Score} is the log probability of the GPT-2, and $\lambda$s are the scaling factors. 
where $P_\text{Whisper}(Y|X)$ is the probability of hypothesis $Y$ given acoustic feature $X$ predicted by Whisper, $P_\text{ilm}(Y)$ and $P_\text{GPT-2}(Y)$ are the internal and external LM probabilities of $Y$, and $\lambda$'s are hyper-parameters to be tuned.

\begin{table*}[t]
    \vspace{-0.3cm}
    \centering
    \caption{WER and R-WER on LibriSpeech test sets, SLURP test set and DSTC2 test set using Whisper and TCPGen, rescored with GPT-2. Test-time biasing list selection followed the description in Sec. \ref{sec:exp}. LM weights tuned on dev sets of each data separately.}
    \vspace{-0.2cm}
    \begin{tabular}{lcccccccc}
    \toprule
    & \multicolumn{2}{c}{Test-clean} & \multicolumn{2}{c}{Test-other} & \multicolumn{2}{c}{SLURP} & \multicolumn{2}{c}{DSTC2} \\
    System & WER $\downarrow$ & R-WER $\downarrow$ & WER $\downarrow$ & R-WER $\downarrow$ & WER $\downarrow$ & R-WER $\downarrow$ & WER $\downarrow$ & R-WER $\downarrow$ \\
    \midrule
     Whisper base.en & 5.2\% & 16.3\% & 10.5\% & 31.4\% & 27.8\% & 72.5\% & 21.1\% & 71.3\% \\
     Whisper base.en + TCPGen & 4.8\% & 12.5\% & 10.1\% & 26.5\% & 25.8\% & 52.3\% & 18.2\% & 41.8\% \\
     Whisper base.en + GPT-2  & 4.9\% & 14.9\% & 9.9\% & 29.4\% & 26.7\% & 67.7\% & 20.8\% & 67.9\% \\
     Whisper base.en + TCPGen + GPT-2 & 4.5\% & 11.7\% & 9.7\% & 25.0\% & 24.8\% & 47.0\% & 16.9\% & 38.7\% \\
    \midrule
     Whisper large    & 4.0\% & 10.4\% & 6.7\% & 20.0\% & 16.7\% & 61.6\% & 17.3\% & 69.1\% \\
     Whisper large + TCPGen & 3.4\% &  8.3\% & 6.3\% & 16.3\% & 15.1\% & 37.1\% & 14.7\% & 33.5\% \\
     Whisper large + GPT-2  & 3.9\% & 10.1\% & 6.6\% & 19.6\% & 16.6\% & 59.3\% & 17.3\% & 68.5\% \\ 
     Whisper large + TCPGen + GPT-2 & \textbf{3.4}\% & \textbf{8.2}\% & \textbf{6.3}\% & \textbf{16.3}\% & \textbf{15.0}\% & \textbf{37.1}\% & \textbf{13.9}\% & \textbf{29.5}\% \\
    \bottomrule
    \end{tabular}
    \vspace{-0.3cm}
    \label{tab:wer1}
\end{table*}

\section{Experimental Setup}
\label{sec:exp}
\subsection{Data and test-time biasing lists}

Experiments were performed on three datasets, including LirbiSpeech clean data, SLURP data and the dialogue state tracking challenge Track 2 (DSTC2). LirbiSpeech represents a fairly generic dataset while the other two are more specific to certain tasks. Descriptions of each data set and the extraction of the biasing list during inference are provided below.

\textbf{LibriSpeech} is an audiobook dataset. The train-clean-100 set was used for training TCPGen and the two test sets were used for evaluation. Following the simulation in \cite{DBRNNT,TCPGen}, a full rare word list containing 200k words was first selected by removing the 5k most frequent words from the LibriSpeech LM training vocabulary. During inference, one biasing list for each utterance was obtained by collecting words that appeared in the full rare word list and adding a certain number of distractors, which, if not specified, was 1000 for LibriSpeech and SLURP.

\textbf{SLURP} \cite{slurp} is a dataset containing single-turn user interactions with a home assistant, annotated with scenarios, actions and entities, where 58 hours of real speech data were used for training. Following \cite{tcpgenslu}, the full biasing list was obtained by extracting words from slot entities that appeared in the error-based biasing list for training. Out-of-training-set words in slot entities were also included, making the full biasing list contain 2.8k words in total. The biasing list for each utterance during inference was organised in the same way as LibriSpeech.

\textbf{DSTC2} is a human-machine task-oriented dialogue dataset where the user-side speech input was used for training and evaluation, which corresponds to around 10 hours of speech for training, validation and testing respectively. The biasing list for inference was organised by collecting ontology words that appeared in the error-based biasing list for training, and, as SLURP, adding out-of-training-set words. The full biasing list contains 380 words due to the small ontology and hence was used as a whole without the need for reference texts.

Note that as the Whisper tokeniser is case sensitive, for each biasing word, a copy of it with the first character capitalised was also included in the biasing list, which doubled the size of each biasing list. The feature extraction, pre-processing and tokenisation followed the Whisper pipeline \cite{whisper}. 

\subsection{Model specifications and evaluation metrics}

Experiments were performed with both the Whisper base.en model trained on English data and the Whisper large trained on multilingual data\footnote{Code for Whisper models: https://github.com/openai/whisper}. The base model contained a 6-block Transformer encoder and decoder with 74M parameters, and the large model contained a 32-block Transformer encoder and decoder with 1550M parameters. TCPGen was trained separately on each data for 30 epochs using an Adam optimiser with a linear tri-state learning rate scheduler. The training was performed on a single A100 GPU. As an example, the training time of the base.en model with TCPGen on LibriSpeech clean-100 was 5 hours. GPT-2 base model was used to rescore 50-best lists without fine-tuning, with $\lambda$'s searched between 0 and 1 on validations sets respectively.

In addition to WER, R-WER \cite{DBRNNT, TCPGen,tcpgenslu} was also used to evaluate system performance. R-WER is the total number of {error} word tokens that belong to the biasing list divided by the total number of word tokens in the test set that belong to the biasing list. Insertion errors were counted in R-WER if the inserted word belonged to the biasing list. Moreover, to keep the performance comparable and compatible with other biasing setups \cite{DBRNNT,TCPGen}, unless indicated, text normalisation was not used for scoring, and a discussion on this is provided in Sec. \ref{sec:ablation}.

\section{Results}
\label{sec:result}

\subsection{Overall Evaluation}

The main results were summarised in Table \ref{tab:wer1}. Compared to the Whisper base.en model, using contextual biasing achieved an average 20\% relative R-WER reduction on test-clean and test-other sets, a 28\% relative R-WER reduction on SLURP and a 42\% relative R-WER reduction on DSTC2. Reductions on domain-specific data were much larger than that on the generic LibriSpeech data, since the test time biasing list of LibriSpeech contained mainly generic words, whereas that for the other two test sets contained domain-specific words such as names of podcasts or restaurants. As biasing word tokens in LibriSpeech and SLURP occupied around 10\% of the total number of word tokens in the test sets, the reduction in overall WER was greater than the error reduction reflected in R-WER, indicating that contextual biasing was also beneficial to unbiased words rather than degrading their performance. This effect was most obvious for the DSTC2 data. Contextual biasing achieved larger improvements with the large multilingual Whisper model, with 40\% relative R-WER reduction on SLURP and over 50\% R-WER reduction on DSTC2.

When GPT-2 was applied to the Whisper base.en model, relative R-WER reductions became slightly smaller for LibriSpeech as GPT-2 served as another generic knowledge source, where the R-WER reduction was retained for other datasets. However, applying GPT-2 to the large model resulted in a much smaller improvement than to the base.en model. As before, larger relative WER and R-WER reductions were observed with the large multilingual model when GPT-2 was used for rescoring. TCPGen achieved 18\% relative R-WER reduction on LibriSpeech test sets, 37\% on SLURP and 57\% on DSTC compared to the Whisper large model with GPT-2 rescoring.

\subsection{Discussion}
\label{sec:ablation}

\begin{figure}[h]
    \centering
    \includegraphics[scale=0.32]{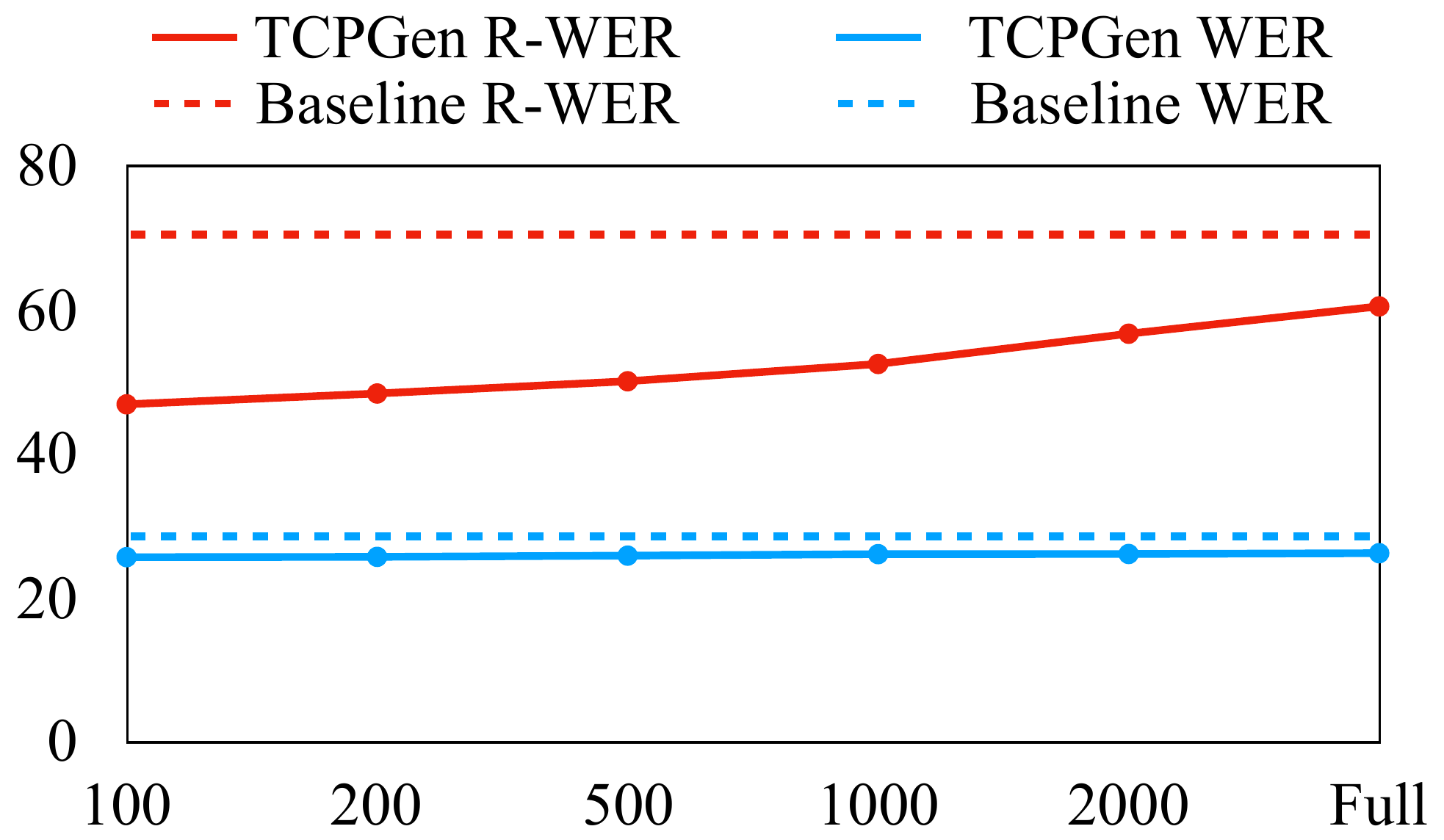}
    \vspace{-0.2cm}
    \caption{WER and R-WER on SLURP using TCPGen and Whisper base.en model. The baseline is base.en without TCPGen. Full had 5.6k biasing words.}
    \label{fig:bsize}
    \vspace{-0.3cm}
\end{figure}

First, the variation of model performance against the \textbf{sizes of the test-time biasing list} was examined, where SLURP was used for this purpose as a non-generic test set, as shown in Fig. \ref{fig:bsize}. Up to the full-sized biasing list of 5.6k biasing words, which achieved 60.5\% R-WER which did not require any knowledge about the reference, TCPGen still outperformed the baseline by a relative 18\% reduction in R-WER.

\begin{table}[t]
    \centering
    \caption{WER on LibriSpeech test-clean using Whisper large model and TCPGen, together with text normalisation. A beam size of 10 was used. }
    \vspace{-0.2cm}
    \begin{tabular}{lccc}
    \toprule
    System & WER $\downarrow$ & R-WER $\downarrow$ \\
    \midrule
     Whisper norm. & 2.5\% & 8.1\% \\
     Whisper + TCPGen norm. & \textbf{2.3}\% & \textbf{7.0}\% \\
    \bottomrule
    \end{tabular}
    \vspace{-0.2cm}
    \label{tab:norm}
\end{table}
Meanwhile, it is necessary to discuss the influence of \textbf{text normalisation} used in the Whisper paper as a post-processing stage for both reference and hypothesis. Results with normalisation were provided in Table \ref{tab:norm} for LibriSpeech test-clean set.

Text normalisation had a similar effect on test-other and SLURP, with a relative R-WER reduction of 25\% on SLURP. It was much less influential to DSTC2 and the relative R-WER improvement remained at 50\%. Nevertheless, text normalisation may obscure the improvements in a biasing word \footnote{For instance, changes in possessive subtractions or British or American spellings that correspond to words in the biasing list}, it was necessary to evaluate the performance of contextual biasing without text normalisation.

% \subsubsection{Error-based biasing list}
\textbf{Using error-based biasing list}: Since the training set for Whisper was unknown, the discrepancy in the task-specific training set may cause degraded performance with a frequency-based biasing list (as used in \cite{DBRNNT, TCPGen}) extracted from it. In fact, for generic datasets such as LibriSpeech train-clean-100, using a frequency-based biasing list by including words with a training set frequency of less than 15, similar R-WER reductions were achieved to the error-based one. However, the error-based biasing list for training achieved a 12\% greater reduction in R-WER compared to the frequency-based one for domain-specific sets, SLURP and DSTC, as the distribution difference to the Whisper training data is much larger.\footnote{Unknown training set also limited word-frequency-based analysis}

% Next, \textbf{the effect of using an error-based biasing list} compared to using a frequency-based biasing list is shown in Table \ref{tab:errorbased}. As before, SLURP was used for this purpose. As the frequency-based biasing list had a much lower average WER than the error-based one, TCPGen was not needed as much by the Whisper model. As a result, using an error-based biasing list for training achieved 12\% additional R-WER reduction compared to the frequency-based one.

% \begin{table}[h]
%     \centering
%     \caption{Effect of the error-based biasing list for training TCPGen evaluated using SLURP data on the Whisper base model. The frequency biasing list contained words that appeared less than 30 times in the training set.}
%     \begin{tabular}{lcc}
%     \toprule
%     Biasing list & WER (\%) & R-WER (\%) \\
%     \midrule
%     Baseline & 27.8\% & 72.5\% \\
%      Frequency-based  & 26.6\% & 59.8\% \\
%      Error-based & \textbf{25.8}\% & \textbf{52.3}\% \\
%     \bottomrule
%     \end{tabular}
%     \label{tab:errorbased}
% \end{table}

% \subsubsection{Performance on unseen words}
Finally, as TCPGen required training on a task-specific set, it is worthwhile examining the \textbf{performance of words that did not belong to that training set}. An OOV WER \cite{tcpgengnn} was used which was measured the same way as R-WER but for words not in the task-specific training set, with results shown in Table \ref{tab:oov}. Whisper with TCPGen biasing achieved 30\% relative OOV WER reduction compared to Whisper alone, which was much higher than that achieved on other biasing words. As a generic dataset, words not covered by the LibriSpeech training set had a high chance of being rare in the Whisper training set and hence TCPGen had a much larger improvement on them. The situation was reversed on SLURP and DSTC as two domain-specific datasets, and words not covered by the training set may be common in the training set of Whisper, especially for DSTC. Therefore, the OOV WER reduction was smaller on those two datasets compared to the R-WER reduction.

\begin{table}[t]
    \centering
    \caption{Effect of TCPGen on words not in the task-specific training set using the Whisper large model, measured by OOV WER defined the same way as R-WER. OOV WER for test-clean and -other was calculated together.}
    \vspace{-0.2cm}
    \begin{tabular}{lccc}
    \toprule
    System & LibriSpeech $\downarrow$ & SLURP $\downarrow$ & DSTC2 $\downarrow$ \\
    \midrule
     Whisper  & 35.0\% & 67.1\% & 15.1\% \\
     Whisper + TCPGen & \textbf{24.5}\% & \textbf{45.3}\% & \textbf{11.7}\% \\
    \bottomrule
    \end{tabular}
    \vspace{-0.3cm}
    \label{tab:oov}
\end{table}

\section{Conclusions}
\label{sec:con}
This paper examined the effectiveness of neural contextual biasing for Whisper in conjunction with GPT-2. Specifically, this paper proposes a method for adapting the TCPGen component to Whisper, along with a dedicated training scheme without the need to fine-tune Whisper. Experiments across three datasets revealed that the proposed method resulted in a large reduction of R-WER when a biasing list of 1000 words was utilized. In particular, \textbf{37\%} and \textbf{57\%} relative R-WER reductions were achieved on SLURP and DSTC2 respectively. The findings also suggested that contextual biasing is more effective when applied to domain-specific data and has the potential to enhance the performance of universal models without sacrificing their capacity for generalization. Though only Whisper and GPT-2 were investigated in this paper, as future work, the biasing scheme can be easily integrated with other universal ASR and LMs, such as Google's USM \cite{usm} if the parameters are open-source.
% \section{Acknowledgements}

% \bibliographystyle{IEEEtran}
% \bibliography{mybib}

\end{document}